\documentclass[10pt,twocolumn,letterpaper]{article}

\usepackage[pagenumbers]{cvpr}

\usepackage[dvipsnames]{xcolor}

\definecolor{cvprblue}{rgb}{0.21,0.49,0.74}
\usepackage[pagebackref,breaklinks,colorlinks,citecolor=cvprblue]{hyperref}
\usepackage{csquotes,amsmath,enumitem,tabularray,fancyhdr,chngcntr}

\newcommand{\arXivfootnote}{
    \fancyhead{}
    \renewcommand{\headrulewidth}{0pt}
    \fancyfoot[LO]{\footnotesize{arXiv preprint.}}
    \fancypagestyle{firstpage}{
      \fancyhead{}
      \renewcommand{\headrulewidth}{0pt}
      \fancyfoot[LO]{\footnotesize{arXiv preprint.}}
    }
    \thispagestyle{firstpage}
}

\title{inkn'hue: Enhancing Manga Colorization\\
from Multiple Priors with Alignment Multi-Encoder VAE
}

\author{Tawin Jiramahapokee\\
Montfort College, Chiang Mai, Thailand\\
{\tt\small tawinj@montfort.ac.th}
}

\begin{document}
\twocolumn[{
\maketitle
\vspace*{-1.5em}
\begin{center}
    \centering
    \captionsetup{type=figure}
    \includegraphics[width=0.99\textwidth]{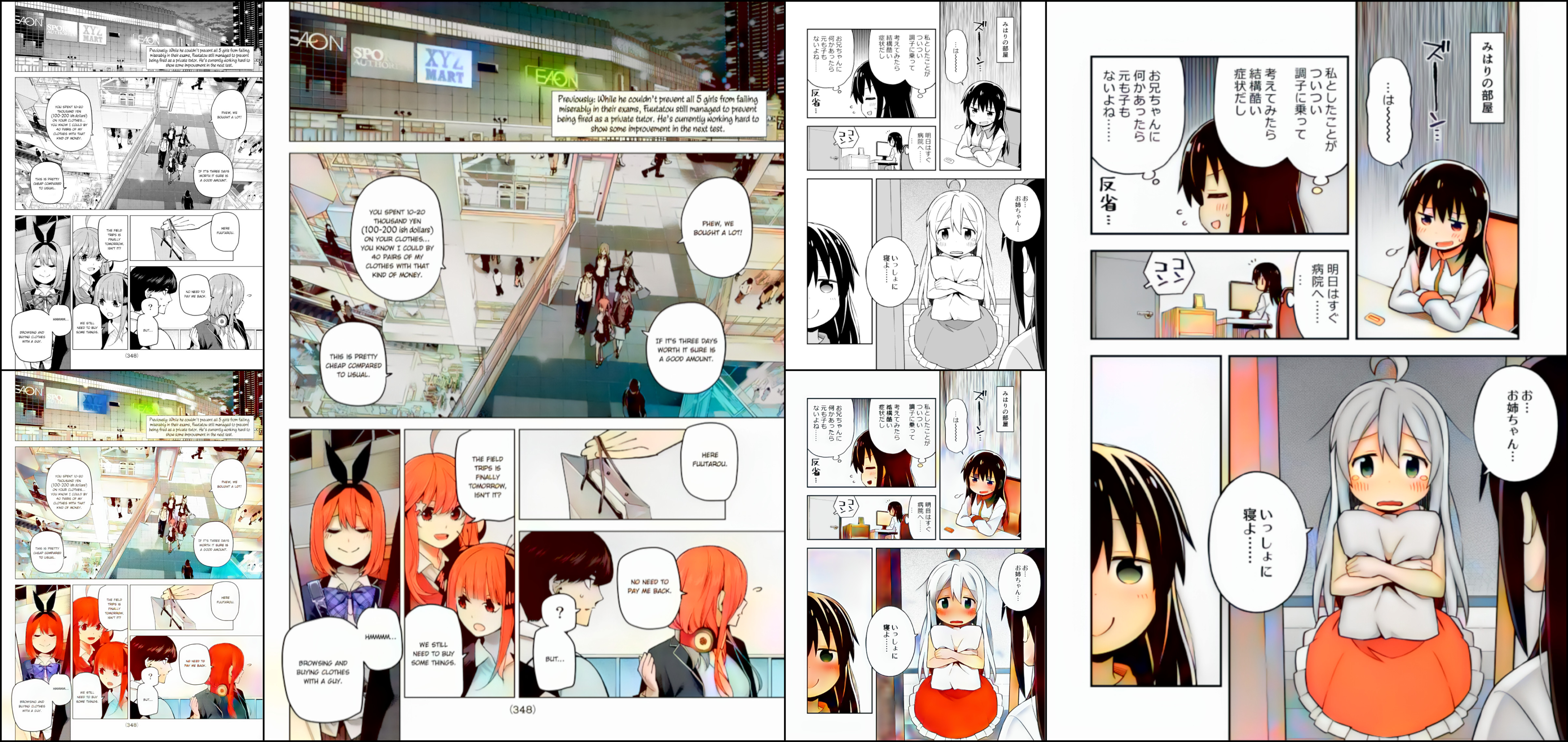}
    \captionof{figure}{\textbf{Examples of original grayscale inputs (top-left) and rough color inputs (bottom-left).} Final colorizations (right) are from our multi-encoder VAE outputs blended with rough color inputs in CIELAB color space ($\lambda_{a*b*}=0.8$)}\label{fig:1}
\end{center}
}]

\arXivfootnote
\begin{abstract}
Manga, a form of Japanese comics and distinct visual storytelling, has captivated readers worldwide. Traditionally presented in black and white, manga's appeal lies in its ability to convey complex narratives and emotions through intricate line art and shading. Yet, the desire to experience manga in vibrant colors has sparked the pursuit of manga colorization, a task of paramount significance for artists. However, existing methods, originally designed for line art and sketches, face challenges when applied to manga. These methods often fall short in achieving the desired results, leading to the need for specialized manga-specific solutions. Existing approaches frequently rely on a single training step or extensive manual artist intervention, which can yield less satisfactory outcomes. To address these challenges, we propose a specialized framework for manga colorization. Leveraging established models for shading and vibrant coloring, our approach aligns both using a multi-encoder VAE. This structured workflow ensures clear and colorful results, with the option to incorporate reference images and manual hints.
\end{abstract}
\section{Introduction}
\label{sec:intro}
Recent work in non-photorealistic colorization primarily focuses on line art and sketch colorization tasks \cite{Zhang2018,Zhang2018style,Zhang2021,Carrillo2023,Ci2018}. While these methods have shown promise in their intended applications, they are not inherently suitable for manga colorization. Implementing them for manga colorization typically necessitates additional steps to address issues like color bleeding and maintaining text clarity.

In the context of manga colorization, some existing solutions are available \cite{Furusawa2017,Xie2020}. However, these solutions also encounter similar challenges, and some may require either a single, extensive training step \cite{Hensman2018} or involve manual efforts from the artist in creating flat colors \cite{Shimizu2021}, which can be a time-consuming process. Moreover, when it comes to using handcrafted algorithms for manga colorization, the results often appear flat and unappealing \cite{Sato2014}.

Our approach draws upon well-established models known for their proficiency in generating high-quality shading results in manga illustrations in addition to decent colors \cite{Golyadkin2021}, as well as models recognized for their ability to produce results with a wide spectrum of vibrant and diverse colors \cite{Zhang2018style,Zhang2021}.

\begin{center}
    \centering
    \captionsetup{type=figure}
    \includegraphics[width=0.46\textwidth]{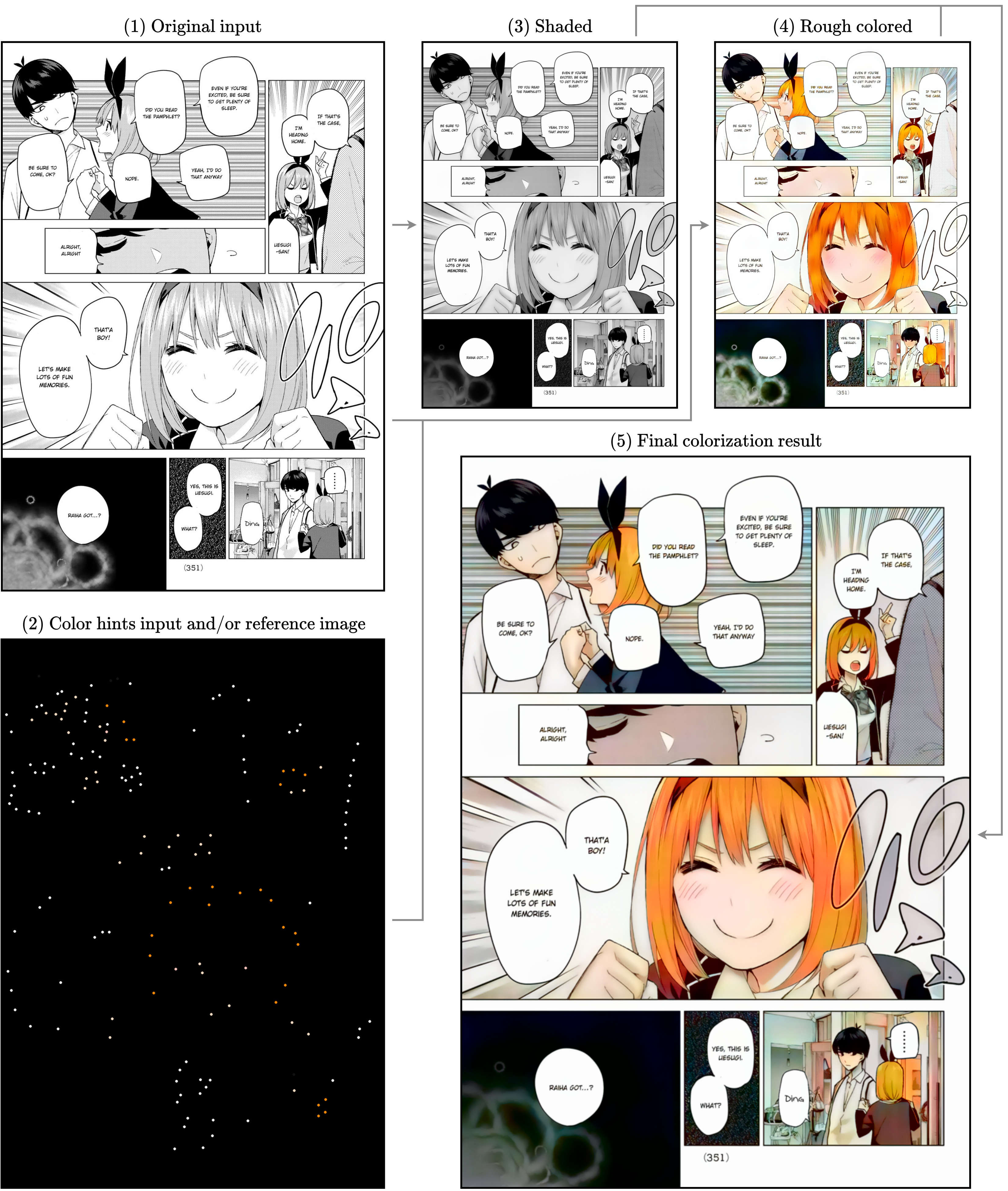}
    \captionof{figure}{\textbf{Overview of the stages of our colorization framework.} Starting with the original image (1), the shading model generates a shaded grayscale version (3). Alongside this, the colorization model produces an initial rough-colored version (4) guided by additional cues provided by user-inputted color hints and/or a reference image (2). The combination model combines both the shaded (3) and rough-colored (4) stages, interpolating colors from the latter to produce the final colorization result (5).}\label{fig:2}
\end{center}

Nevertheless, the straightforward approach of merging the shaded grayscale stage with the preliminary rough-colored stage via value blending within the CIELAB color space \cite{Zeyen2018} does not consistently yield desirable outcomes. In specific situations, the initial rough-colored results exhibit peculiar anomalies, characterized by unconventional color patches and inconsistencies within the painted regions. These occurrences highlight the limitations of a one-size-fits-all methodology. Therefore, a more sophisticated strategy is needed, as explored in our proposed framework.

Recognizing the need for a more nuanced solution, we introduce an additional network into our framework. This network is specifically designed to address the deficiencies within the initial rough-colored stages. Furthermore, it undertakes the crucial task of harmonizing the color palette across the entire image, ensuring a cohesive and visually appealing end result. The integration of this component significantly enhances our approach, addressing any inconsistencies and peculiarities that may arise during the early phases of the colorization process. Consequently, our comprehensive framework is aimed at not only streamlining the colorization process but also enhancing the overall quality and aesthetic appeal of the final output.

By aligning the results from both the shading and colorization models using a multi-encoder VAE \cite{Kingma2014}, our approach provides a structured workflow for producing colorful and plausible results with clear and readable text. Additionally, it offers the flexibility to incorporate reference images and manual hints as sources for the initial colorization. This approach not only enhances the accuracy and fidelity of the colorization process but also empowers artists and users to exert greater creative control. Additionally, this inclusive approach caters to a diverse array of colorization scenarios, ensuring that the final results align closely with the artist's intent and vision.

We validate the influence of our method by showcasing its abilities in producing colorization results with detailed shading and visually appealing colors from a comparatively effortless process of inputting color hints and/or reference images. Additionally, to assess the perceptual improvements offered by our approach, we conduct a user preference test, comparing our refined post-processed results to preliminary colorized images generated by earlier models \cite{Zhang2018}. Our primary contribution is to present a framework for delivering high-quality manga colorization results from color hints and/or reference images. Our secondary contribution is to demonstrate the effectiveness of a multi-encoder VAE architecture in dealing with inconsistencies observed in the generated samples produced by previous methods.
\section{Related work}
\label{sec:related}

\subsection{Photorealistic colorization}
\noindent\textbf{Pix2Pix \cite{Isola2017}} represents a significant milestone in conditional image-to-image translation, employing a cGAN \cite{Mirza2014} architecture. For colorization, it was aimed to tackle the task of generating full-color images from grayscale inputs \cite{KumarSingh2023}. While Pix2Pix was successful in various domain translation tasks, it faced challenges when applied to tasks like line art and sketch colorization \cite{Rodrigues2022,Seo2021}, akin to manga colorization. In these scenarios, the model struggled to accurately infer painting regions and appropriate colors due to the less explicit shading information inherent in such inputs.

\twocolumn[{
\begin{center}
    \centering
    \captionsetup{type=figure}
    \includegraphics[width=\textwidth]{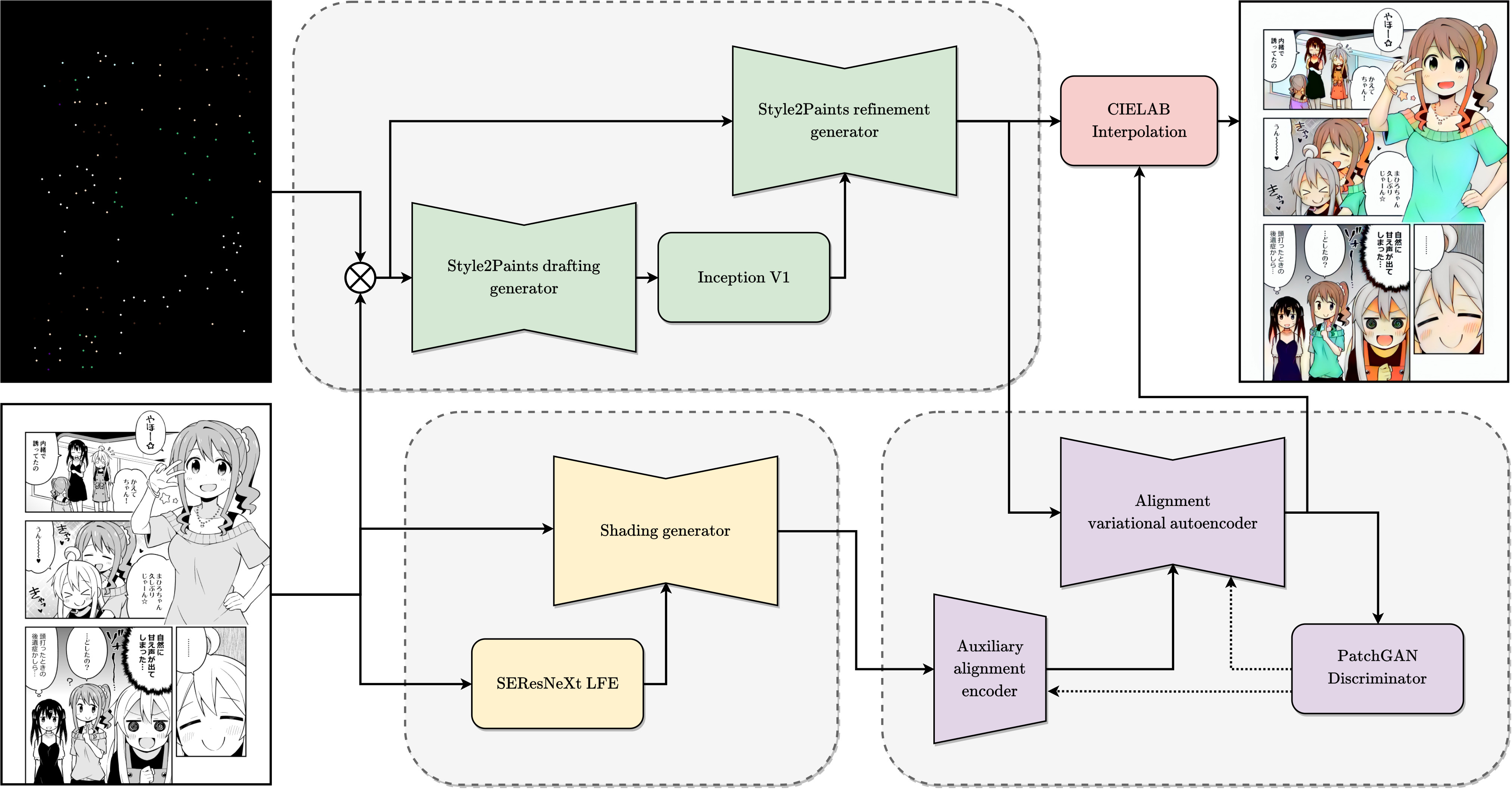}
    \captionof{figure}{\textbf{Expanded overview of the framework.} Our pipeline utilizes trained parameters from related works including Style2Paints \cite{Zhang2018} (shown in green), manga-colorization-v2 \cite{Golyadkin2021} (shown as \textquote{Shading generator}), and Tag2Pix \cite{Kim2019} extractor (shown as \textquote{SEResNeXt LFE} (Local Feature Extractor) \cite{Hu2020}). The framework aligns results from the shading generator (shaded grayscale) and Style2Paints (rough-colored) using an alignment variational autoencoder and an auxiliary alignment encoder (shown in violet). The input consists of the original manga pages (bottom-leftmost), along with the color hints and/or reference images (top-leftmost) that are to be used as local and global color hints, respectively. The outputs from the last-stage model are then interpolated with the rough-colored outputs (shown in red) based on a user-defined interpolation value $\lambda_{a*b*}$ to produce the most appealing final colorized results (top-rightmost).}\label{fig:3}
\end{center}
}]

\noindent\textbf{Latent diffusion-based image colorization \cite{Liu2023}} exploits color prior knowledge from text-to-image latent diffusion models \cite{Rombach2022} through piggybacking off them and seperately training a diffusion guider model. Additionally, it introduces the concept of a lightness-aware VQVAE \cite{Esser2021} to incorporate grayscale information, ensuring pixel-aligned results. This approach offers remarkable flexibility for image colorization, accommodating text prompting and the potential incorporation of color hints (though not in the cited work), especially with recent conditional control techniques \cite{Zhang2023}. However, training these diffusion models demands a substantial dataset of paired captioned images, a challenge that proves particularly daunting in the context of manga colorization.

\subsection{Line art and sketch colorization}
\noindent\textbf{Style Transfer for Anime Sketches \cite{Zhang2018style}} employs a Residual U-Net \cite{Ronneberger2015}, AC-GAN \cite{Odena2017}, and \textquote{Guide Decoders} to accomplish style transfer from colorized images to sketches. While this approach is effective for style transfer tasks, it falls short of addressing the challenges presented by manga colorization. The complexity arises from the fact that manga pages often comprise multiple panels, each potentially characterized by a distinct artstyle. Consequently, achieving a globally consistent style through style transfer from a single image alone is not a feasable solution.

\noindent\textbf{Line art colorization from color hints \cite{Ci2018,Zhang2018,Carrillo2023,Zhang2021,Silva2019}} mitigates the ambiguity of automatic line art colorization by integrating user-provided color hints, such as hint points or scribbles, in conjunction with the input line art. Several methods have been proposed to transform the two inputs into properly colorized results. These include cGAN-based methods \cite{Ci2018,Silva2019}, separating the task into a drafting and a refinement stage \cite{Zhang2018}. More recently, a method based on the split filling mechanism \cite{Zhang2021}, along with a DDPM-based method (denoising diffusion probabilitic model) \cite{Carrillo2023,Nichol2021,Ho2020} were introduced. While often achieving great results for sketch and line art inputs, when applied to the task of manga colorization, these models suffer from the specific challenges as follows: 
\begin{enumerate}
    \item Color bleeding occurs between regions that are hard to distinguish from each other.
    \item Text areas experience color filling and alterations in line thickness, affecting text clarity.
    \item Partial degradation of original handstrokes and distinctive lines are observed.
\end{enumerate}

\subsection{Manga colorization}
\noindent\textbf{Manga colorization from a single training image \cite{Hensman2018}} utilizes cGAN training on a small dataset along with segmentation \cite{Zhang2009} and color-correction for automatic manga colorization. However, relying exclusively on a pretrained cGAN model for colorization presents major limitations. Target images with colors and styles differing from those in the training dataset cannot be processed using the same parameters, necessitating model retraining. Furthermore, the obtained results often exhibit desaturation and lack color variety, even after post-processing, and there are no further mechanisms of control over the final results.

\noindent\textbf{Manga filling style conversion \cite{Xie2020}} performs conversions between western-style color comics and manga by initially mapping both to a common intermediate representation. This transformation involves the use of a \textquote{screentone VAE} for manga and a bidirectional translation model comprising a 7-level U-Net \cite{Ronneberger2015} for color comics. The model falls into the category of automatic colorization, lacking additional control mechanisms for manual intervention. Thus, its utility in a manga colorization task is limited. It is important to note that the model's generated color comics results tend to be unrealistic and visually unappealing.

\noindent\textbf{Manga colorization from flat colored inputs \cite{Shimizu2021}} accepts a pair of screen tone and flat colored images as prior inputs to create properly shaded colorized results. The method utilizes the screen tone images to provide essential shadow and lightness information, crucial for achieving high-quality results. Although traditional black and white manga may not inherently contain sufficient shadow and lightness details, the method can still implicitly infer shading by combining the input image pairs. However, creating a flat colored image from scratch as an input for the model can be a time-consuming and challenging task, especially for non-artists.

\noindent\textbf{Comicolorization \cite{Furusawa2017}} is a CNN-based approach \cite{Saxena2022} designed for the colorization of segmented manga panels. The process begins with the extraction of individual manga panels \cite{Ishii2009}, followed by a style transfer step using a reference image. Afterward, a user interactive revision step is employed, and finally, the panels are recombined during a layout restoration phase. This method requires multiple reference images per page, making it less practical. Additionally, for manga pages with non-standard layouts, separating the panels may not be feasible, which restricts its use to a fixed set of supported manga. Lastly, the colorized results exhibit the issues of color bleeding and large uncolored areas.

\noindent\textbf{Semi-automatic manga colorization with color hints \cite{Golyadkin2021}} has demonstrated the most promising manga colorization results. The architecture comprises a SEResNeXt U-Net \cite{Hu2020,Ronneberger2015} with a local feature extractor, and a discriminator containing convolutional and SEResNeXt blocks. Results from the model exhibit realistic and high-quality shading, although the model tends to overuse the \textquote{white color penalty}, leading it to aggressively paint in regions where a color closer to white is expected. Additionally, the color hint mechanism does not support the incorporation of global hints from reference images, making it challenging to achieve the desired tone. Support for colorization with color hints was dropped in the improved version of the model, manga-colorization-v2\footnote{https://github.com/qweasdd/manga-colorization-v2\label{manga_related}}. Furthermore, the FFDNet \cite{Zhang2018_ffd} applied in the process originally intended to remove noise present in the images often introduces blurriness to the images and text.

\section{Method}
\label{sec:method}

\noindent\textbf{Overview.} Our objective is to develop a framework from prior models while deriving and training a recombination model. We define the input to comprise of the original black and white manga page $I_{bw}\in\mathbb{R}^{h \times w}$, user-defined color hints $I_{h}\in\mathbb{R}^{h \times w \times 3}$, and optional reference image $I_{ref}\in\mathbb{R}^{h \times w \times 3}$. The height and width of the images are denoted as $h$ and $w$, respectively. The output is defined as a colorized image $Y\in\mathbb{R}^{h \times w \times 3}$. Our framework can be divided into shading, rough colorization, recombination, and post-processing stages. The chosen prior models are Style2PaintsV4.5\footnote{https://github.com/lllyasviel/style2paints} \cite{Zhang2018} and manga-colorization-v2\footref{manga_related} \cite{Golyadkin2021}. We initilize our multi-encoder VAE-GAN \cite{Larsen2016} training with model parameters from kl-f8 \cite{Rombach2022}. Additionally, parameters from Tag2Pix \cite{Kim2019} are used within the shading model.

\subsection{Prior models}
\noindent\textbf{Selection of prior models.} Our choice of prior models was made after consideration of their alignment with our specific criteria. The models discussed (see Section \ref{sec:related}) had a common issue of imprecision in colorization. However, one exception was identified, namely, \textquote{Semi-automatic Manga Colorization Using Conditional Adversarial Networks} \cite{Golyadkin2021}, which exhibited exceptional shading quality. Thus, we decided to integrate this model into our framework as the shading model. Additionally, for the task of generating bold and realistic colors with a high degree of user control, Style2Paints \cite{Zhang2018} was the standout choice, and it was adopted as our rough colorization model.

\noindent\textbf{Improvements.} While our selected models demonstrated notable proficiency, we also identified some minor issues with the chosen rough colorization model. In particular, it occasionally exhibited color bleeding when provided with a reference image while having insufficient color hints. The model also occasionally inadvertently colorized text regions, and sometimes filled spaces between manga panels with color. To address these inconsistencies, we determined that our recombination model should include enhancements specifically tailored to mitigate these issues.

\twocolumn[{
\begin{center}
    \centering
    \captionsetup{type=figure}
    \includegraphics[width=\textwidth]{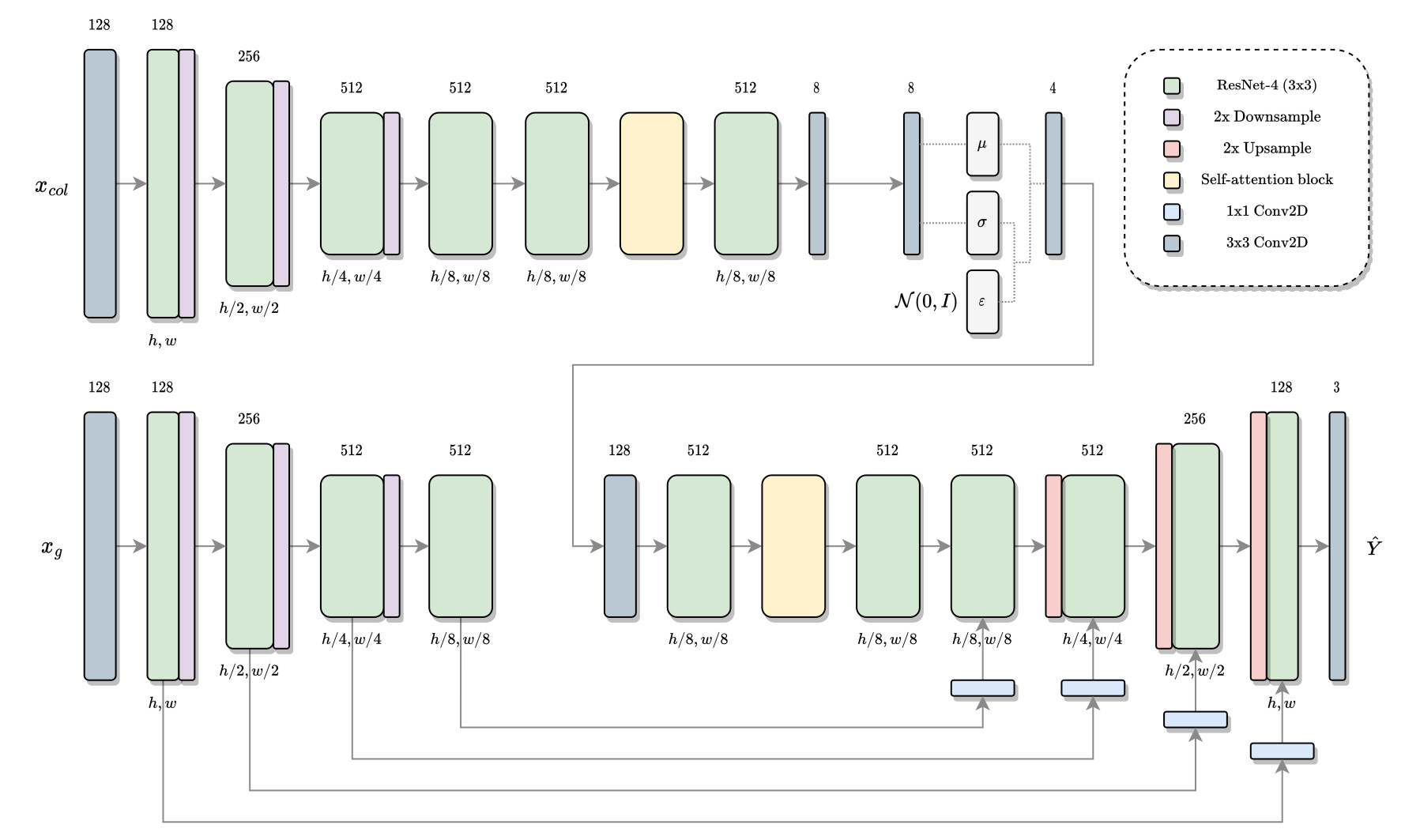}
    \captionof{figure}{\textbf{Architectural diagram of the alignment multi-encoder variational autoencoder.} The number of feature dimensions of the output are depicted at the top, while the input resolutions are indicated at the bottom of each subnetwork block.}\label{fig:4}
\end{center}
}]

\subsection{Network architecture}
\noindent\textbf{Prior models.} The architecture of the Style2Paints rough colorization model remains unaltered. The model accepts $I_{bw}$ and $I_{h}$ as inputs, with the option to include $I_{ref}$. In cases where $I_{ref}$ is not provided, the model defaults to using a randomly selected color palette for colorization. Its output consists of the rough-colored intermediate results $x_{col} \in \mathbb{R}^{h \times w \times 3}$. The shading model takes $I_{bw}$ as input and generates shaded grayscale intermediate results denoted as $x_{g} \in \mathbb{R}^{h \times w}$. We introduced a specific adjustment by eliminating the FFDNet \cite{Zhang2018_ffd} denoising steps from the shading model, while the remaining architectural components of the model have been retained without modifications.

\noindent\textbf{Generator.} The generator incorporates a Variational Autoencoder (VAE) structure based on the kl-f8 model \cite{Rombach2022}. The VAE comprises an encoder $\mathcal{E}$ designed to accept the $x_{col}$ input and convert it into a latent distribution $q_\mathcal{E}(z | x_{col})$. During both training and inference, the VAE samples from this latent distribution, with the resulting sample denoted as $z$. An auxiliary encoder, architecturally similar to the main encoder but with the middle blocks removed, further augments the decoder. Denoted as $\mathcal{E}_g$, it provides additional grayscale information to the decoder. The auxiliary encoder takes $x_g$ as input and shortcuts the decoder towards the target output distribution by connecting to the decoder's upsampling blocks through skip connections \cite{He2016} via 1x1 convolutional blocks from each of the auxiliary encoder's downsampling block. The decoder, labeled as $\mathcal{D}$, maintains the same structure as found in the kl-f8 model, but it features skip connections from the auxiliary encoder $\mathcal{E}_g$. Each of these residuals is added to the output of the corresponding upsampling blocks of the decoder. Additionally, the decoder uses the latent sample $z$ to generate the recombined intermediate result $\hat{Y}$. During training, $x_g$ is synthesized from the colored training dataset by applying a grayscale transform to the colored images. $x_{col}$, on the other hand, is synthesized using solely $I_{bw}$ and $I_{ref}$ as inputs for the Style2Paints model, without the inclusion of color hints $I_h$. This deliberate degradation of $x_{col}$ during training serves the purpose of making the model robust to poorly colorized rough-colored inputs, akin to DVAEs \cite{Im2017}.

\noindent\textbf{Discriminator.} The model incorporates a PatchGAN discriminator $\mathcal{D}_\psi$ \cite{Rombach2022,Li2016} with a patch size of $64$. This discriminator is employed for the model's adversarial objective. During training, the discriminator provides updates to the auxiliary encoder $\mathcal{E}_g$ and the decoder $\mathcal{D}$. The main encoder $\mathcal{E}$ remains frozen during training. This decision was based on the reasoning that there is no necessity to alter or adapt the distribution $q_{\mathcal{E}}(z | x_{col})$ during the training process.

\noindent\textbf{Parameter initialization.} The parameters for the PatchGAN discriminator $\mathcal{D}_\psi$, encoder $\mathcal{E}$, auxiliary encoder $\mathcal{E}_g$, and decoder $\mathcal{D}$ are initialized from the kl-f8 model \cite{Rombach2022}. The 1x1 convolutions are initialized with parameters set to zero, following the approach taken by the \textquote{zero-convolution} layers in \textquote{Adding Conditional Control to Text-to-Image Diffusion Models} \cite{Zhang2023}. Furthermore, the parameters of the encoder $\mathcal{E}$ were frozen during training. The parameters of the prior models remain unchanged and are not trained further.

\noindent\textbf{Training pipeline.} The training process can be summarized through the following steps:
\begin{enumerate}
    \item Given $I_{col}$, apply grayscale transformations to create $x_g$.
    \item Apply the Style2Paints rough colorization stage with $x_g$ as $I_{bw}$ and $I_{col}$ as $I_{ref}$, without $I_h$, to generate $x_{col}$.
    \item Run a forward pass on the generator, using $x_{col}$ and $x_g$, to generate $\hat{Y}$.
    \item Calculate the loss from $\hat{Y}$ and $I_{col}$.
    \item Update parameters of the generator and discriminator in two separate steps.
\end{enumerate}
\noindent\textbf{Inference pipeline.} The inference process are as follows:
\begin{enumerate}
    \item Take in $I_{bw}$, $I_h$, and optionally $I_{ref}$ as inputs.
    \item Apply the Style2Paints rough colorization stage with $I_{bw}$, $I_h$, and optionally $I_{ref}$, to generate $x_{col}$.
    \item Apply the shading stage with $I_{bw}$, to generate $x_g$.
    \item Run a forward pass on the generator, using $x_{col}$ and $x_g$, to generate $\hat{Y}$.
    \item Split $\hat{Y}$ and $x_{col}$ into CIELAB $L*$, $a*$, and $b*$ channels
    \item Use the $L*$ channel from $\hat{Y}$ as the $L*$ channel for $Y$
    \item Interpolate the $a*$ and $b*$ channels from $\hat{Y}$ and $x_{col}$ with 
    \begin{equation}
        a* = \hat{Y}^{a*}\cdot(1-\lambda_{a*b*}) + x_{col}^{a*}\cdot\lambda_{a*b*}
    \end{equation}
    \begin{equation}
        b* = \hat{Y}^{b*}\cdot(1-\lambda_{a*b*}) + x_{col}^{b*}\cdot\lambda_{a*b*}
    \end{equation}
    \item Merge the interpolated $a*$ and $b*$ channels into $Y$
\end{enumerate}

\subsection{Loss}
\noindent\textbf{Overview.} In our approach, we compute the loss using a combination of the \textit{L1 Loss}, \textit{Perception Loss}, and \textit{Adversarial Loss}. Our loss is calculated based on $Y$ and $\hat{Y}$, which differ from the typical reconstruction loss employed in traditional variational autoencoders. Additionally, we don't require the \textit{KL Loss} term in our overall loss function because the encoder $\mathcal{E}$ is frozen during training, resulting in a fixed latent distribution. Common equations for all loss terms are the following:
\begin{equation}
    \varepsilon \sim \mathcal{N}(0, I)
\end{equation}
\begin{equation}
    z \sim q_\mathcal{E}(z | x_{col}) = \mathcal{N}(z; \mathcal{E}_\mu, \mathcal{E}_{\sigma^2})
\end{equation}
\begin{equation}
    \mathcal{E}(x_{col}) = \mathcal{E}_\mu(x_{col}) + \mathcal{E}_\sigma(x_{col}) \cdot \varepsilon
\end{equation}
\begin{equation}
    \hat{Y} = \mathcal{D}(\mathcal{E}(x_{col}), \mathcal{E}_g(x_g))
\end{equation}

\noindent\textbf{L1 Loss.} Pixel-wise absolute difference between $Y$ and $\hat{Y}$.
\begin{equation}
    \mathcal{L}_1 = ||Y - \hat{Y}||_1
\end{equation}

\noindent\textbf{Perception Loss.} Computed using the LPIPS metric \cite{Zhang2018_lpips} based on a pretrained VGG-16 \cite{Simonyan2015}. It emphasizes the perceptual similarity and likeness of higher-level visual features between $Y$ and $\hat{Y}$. The hyperparameter used during training is $\lambda_p = 1$.
\begin{equation}
    \mathcal{L}_p = \lambda_p \cdot \text{LPIPS}(Y, \hat{Y})
\end{equation}

\noindent\textbf{Adversarial Loss.} Given global step $s_g$, the adversarial term starts affecting the overall loss when $s_g \geq \lambda_{\psi_{start}}$.
\begin{equation} 
    \lambda_{\psi_f} = 
    \begin{cases}
        1, & \text{if } s_g \geq \lambda_{\psi_{start}} \\
        0, & \text{otherwise}
    \end{cases}
\end{equation}
Adaptive weight $\lambda_{\psi_a}$ is computed based on the gradient of the discriminator input w.r.t. the last decoder layer, denoted $\nabla_{\mathcal{D}_L}[\cdot]$. $\mathcal{D}_\psi$ loss is based on hinge loss \cite{Lim2017}. The hyperparameters used during training are $\lambda_\psi = 0.5$ and $\delta = 10^{-4}$.
\begin{equation} 
    \mathcal{L}_{adv}^{\mathcal{E}_g, \mathcal{D}} = -\lambda_{\psi_a}\lambda_{\psi_f}\mathcal{D}_\psi(\hat{Y})
\end{equation}
\begin{equation} 
    \mathcal{L}_{adv}^{\mathcal{D}_\psi} = 0.5\cdot\lambda_{\psi_f}(\Gamma(1 - \mathcal{D}_\psi(Y))+\Gamma(1 + \mathcal{D}_\psi(\hat{Y})))
\end{equation}
\begin{equation} 
    \lambda_{\psi_a} = \frac{\lambda_\psi\nabla_{\mathcal{D}_L}[\mathcal{L}_1+\mathcal{L}_p]}{\nabla_{\mathcal{D}_L}[-\mathcal{D}_\psi(\hat{Y})] + \delta}
\end{equation}
\begin{equation} 
    \Gamma(x) = \text{ReLU}(x)
\end{equation}

\noindent\textbf{Combined Loss.} The combined loss $\mathcal{L}$ is then computed.
\begin{equation}
    \mathcal{L} = \min_{\mathcal{E}_g, \mathcal{D}}\max_\psi(\mathbb{E}_{Y,\hat{Y}}[\mathcal{L}_1 + \mathcal{L}_p + \mathcal{L}_{adv}^{\mathcal{E}_g, \mathcal{D}} - \mathcal{L}_{adv}^{\mathcal{D}_\psi}])
\end{equation}

\subsection{Post-processing}
\noindent\textbf{CIELAB interpolation.} Slight unintended deviations from the original rough-colored inputs $x_{col}$ in areas containing large patches of similar colors may occasionally occur in the generator outputs $\hat{Y}$. To address this issue, we employ CIELAB interpolation \cite{Zeyen2018} as a post-processing technique. CIELAB is a color space characterized by three channels: $L*$ (luminance), $a*$ (green-magenta), and $b*$ (blue-yellow). The CIELAB color space was chosen for interpolation because it is close to being perceptually uniform. Within this interpolation process, our primary focus is on the $a*$ and $b*$ channels, which carry chromatic information, while the $L*$ channel is kept the same as from $\hat{Y}$. We interpolate the values in these channels between two images: $\hat{Y}$ (the generator's output) and $x_{col}$ (the rough-colored input). The extent of adjustment is controlled by a user-defined parameter $\lambda_{a*b*}$. This parameter allows users to tailor the balance of colors between the rough-colored version and the generator's output based on their artistic preferences. In essence, post-processing ensures that the final images align with the user's intent and individual stylistic preferences.

\twocolumn[{
\begin{center}
    \centering
    \captionsetup{type=figure}
    \includegraphics[width=\textwidth]{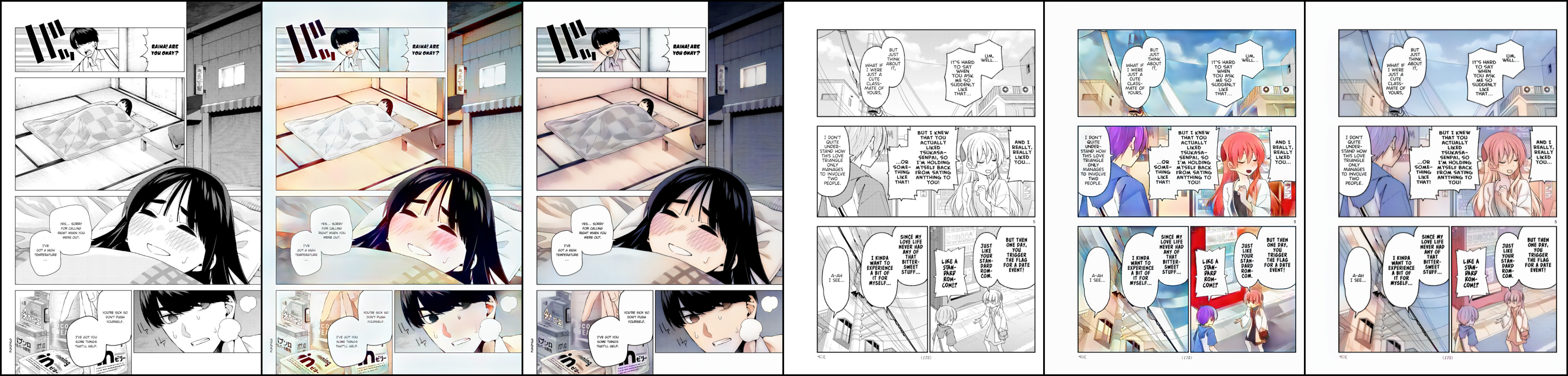}
    \scriptsize{\begin{tblr}{colsep = 0pt, colspec = {*{6}{X[c]}}}
        B\&W input ($I_{bw}$) & Style2Paints ($x_{col}$) \cite{Zhang2018} & Raw generator output ($\hat{Y}$) & B\&W input ($I_{bw}$) & Style2Paints ($x_{col}$) \cite{Zhang2018} & Raw generator output ($\hat{Y}$)
    \end{tblr}}
    \vspace*{-2.0em}
    \captionof{figure}{\textbf{Significance of the generator.} Detail lost from the Style2Paints process are restored, and more accurrate shading is achieved.}\label{fig:5}
\end{center}

\begin{center}
    \centering
    \captionsetup{type=figure}
    \includegraphics[width=\textwidth]{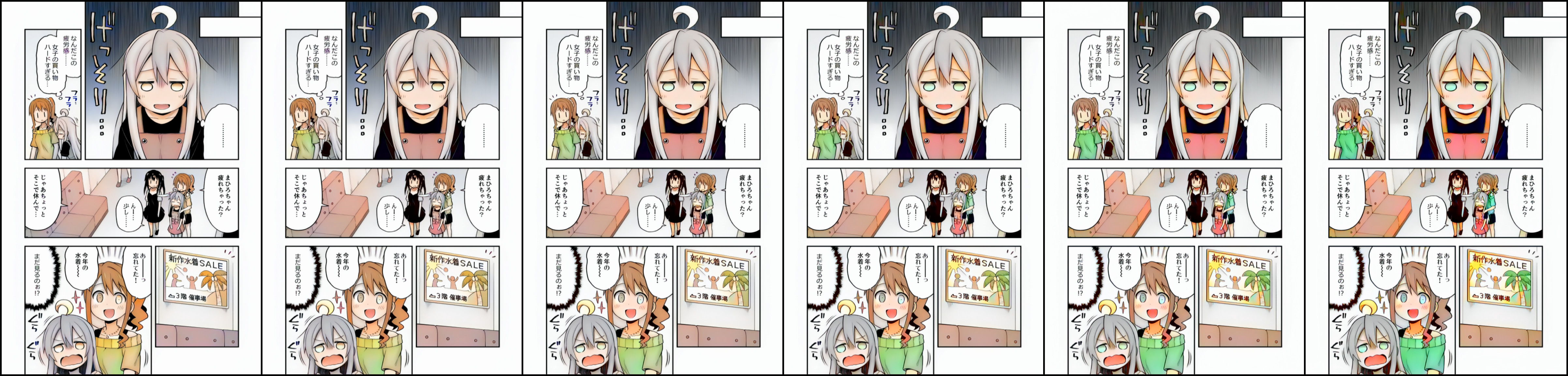}
    \scriptsize{\begin{tblr}{colsep = 0pt, colspec = {*{6}{X[c]}}}
        $\lambda_{a*b*} = 0.0$ & $\lambda_{a*b*} = 0.2$ & $\lambda_{a*b*} = 0.4$ & $\lambda_{a*b*} = 0.6$ & $\lambda_{a*b*} = 0.8$ & $\lambda_{a*b*} = 1.0$
    \end{tblr}}
    \vspace*{-2.0em}
    \captionof{figure}{\textbf{Significance of post-processing.} The generator may desaturate or overcorrect for inaccurate colors without post-processing.}\label{fig:6}
\end{center}
}]
\section{Experiments}
\label{sec:experiments}

\subsection{Experimental setting}
\noindent\textbf{Dataset.} Publicly available datasets for manga colorization were not found in existing literature and related work. Thus, we compiled our training dataset from the Danbooru2021 dataset \cite{Anonymous2022}, Pixiv \cite{pixiv2023}, and MangaDex \cite{mangadex2023}, totaling 58k images. These images feature manga pages containing fully or predominantly colored elements. We manually created color hints of unseen samples for evaluation.

\noindent\textbf{Implementation details.} The training images were resized for the shortest side to have a length of $512$, then random cropped to $256 \times 256$. We implemented the model using PyTorch and Hugging Face Accelerate, running the bfloat16 mixed precision training \cite{Narang2018} for 290k steps and 20 epochs on a single NVIDIA L4 GPU with the AdamW optimizer \cite{Loshchilov2019} using $lr = 4.5\times10^{-6}$, $\beta_1 = 0.9$, $\beta_2 = 0.5$. Other hyperparameters were $\lambda_p = 1$, $\lambda_{\psi_{start}} = 10001$, $\lambda_\psi = 0.5$, $\delta = 10^{-4}$, and a batch size of $4$. Pretrained model parameters were used for kl-f8 \cite{Esser2021}, \textquote{SEResNeXt LFE} \cite{Kim2019}, and the shading model \cite{Golyadkin2021}. Results obtained from Style2Paints were generated using official Style2PaintsV4.5 binaries\footnote{https://github.com/lllyasviel/style2paints}. Comparisons to \cite{Golyadkin2021} are based on manga-colorization-v2\footnote{https://github.com/qweasdd/manga-colorization-v2\label{manga_related2}} which does not support user-guided colorization.

\subsection{Qualitative results}
\noindent\textbf{Line art restoration and shading.} The generator is shown to effectively restore line art details lost during the Style2Paints colorization process (see Fig. \ref{fig:5}, Fig. \ref{fig:7}), which has importance in ensuring small features in the image remain sharp and clear, along with keeping text and fonts readable. Furthermore, the generator produces outputs with higher overall quality and realism of shaded details when compared to Style2Paints rough-colored inputs.

\noindent\textbf{Colors and outliers.} The generator is able to correct color outliers and inconsistencies from the rough-colored stage. However, the initial generator outputs also tend to exhibit desaturation. The subsequent post-processing step is used to ensure final results that are colorful and visually consistent to the surrounding context (see Fig. \ref{fig:6}).

\noindent\textbf{Comparison with baselines.} Results are controllable via color hints and reference images. manga-colorization-v2\footref{manga_related2}, the most recent iteration of the popular manga colorization model, does not support user-guidance. Our model achieves more uniform and truthful colorization results due to being based on a larger line art colorization model for the rough-colorization stage. Our model's inference latency of $\sim$10 seconds on a mid-range laptop NVIDIA GPU results in little additional time cost over colorization with only Style2Paints. Adding color hints take approximately 3-5 minutes with the use of reference images of similar tone.

\twocolumn[{
\begin{center}
    \centering
    \captionsetup{type=figure}
    \includegraphics[width=\textwidth]{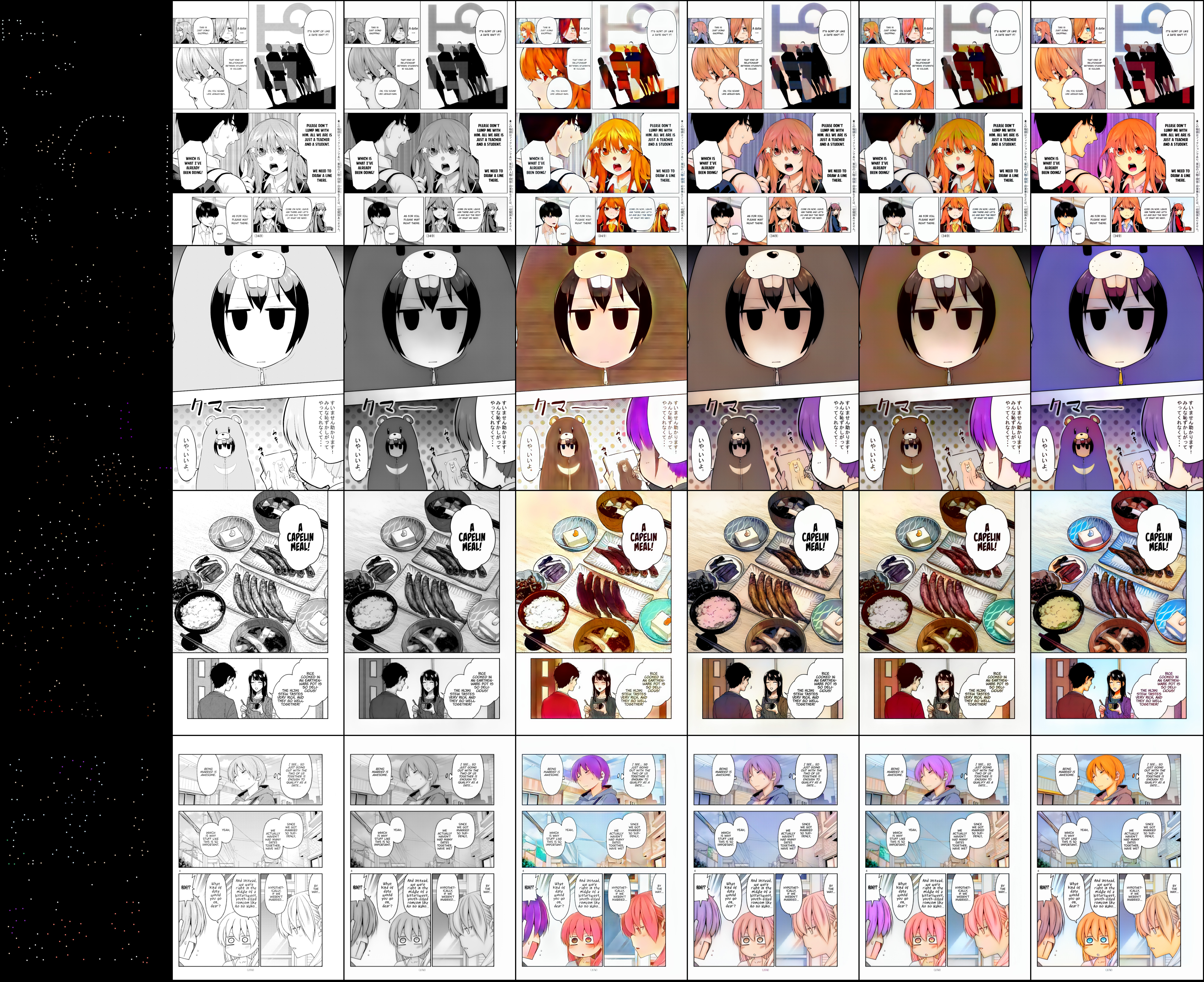}
    \scriptsize{\begin{tblr}{colsep = 0pt, colspec = {*{7}{X[c]}}}
        Color hints ($I_h$) & B\&W input ($I_{bw}$) & Shaded ($x_g$) & Rough-colored ($x_{col}$) & Ours ($\lambda_{a*b*} = 0.0$) & Ours ($\lambda_{a*b*} = 0.8$) & Golyadkin (w/o hint) \cite{Golyadkin2021}
    \end{tblr}}
    \vspace*{-2.0em}
    \captionof{figure}{\textbf{Qualitative comparison.} Additional outputs from each model stage and comparisons to manga-colorization-v2 are shown.}\label{fig:7}
\end{center}
}]

\subsection{Ablation study}
\noindent\textbf{User preference over rough-colored prior.} We conducted qualitative ablation studies to demonstrate the effectiveness of our generator. A user study comparing 26 samples of Style2Paints prior and our post-processed results was done with 31 participants. We provided each participant with results from the two stages in random order and asked them to select the best result with basis in consistency, shading,  clarity, and plausibility of colors. Majority of the users preferred the post-processed over Style2Paints prior results.

\begin{center}
    \centering
    \captionsetup{type=table}
    \vspace*{-1.0em}
    \footnotesize{
    \begin{tblr}{colsep = 0pt, colspec = @{}X[c]X[c]X[c]@{}}
        Model stage         & User preference (\%) $\uparrow$  & Total time cost $\downarrow$ \\
        \hline
        Post-processed      & \textbf{73.08}                   & $\sim$5min \\
        Rough-colored only  & 26.92                            & $\sim$5min \\      
        \hline
    \end{tblr}}
    \vspace*{-0.5em}
    \captionof{table}{\textbf{User preferential alignment} on different model stages.}\label{tab:1}
\end{center}
\section{Conclusion}
\label{sec:conclusion}

\noindent\raggedbottom In this work, we presented a complete approach to user-guided manga colorization that addresses the limitations inherent in existing methods. Our model builds upon the robust colorization capabilities of Style2Paints, along with its flexibility in user guidance, and the shading strengths of manga-colorization-v2. We leverage a multi-encoder VAE to correct for inconsistent color regions of the prior inputs and use it to combine the shading and rough-colored stages. We utilize CIELAB interpolation to improve the color saturation and truthfulness of the final manga colorization output from our generator model, leading to visually pleasing results with remarkable color consistency, high-quality line art, and detailed shading. Our framework provides effective and fast manga colorization in under 3-5 minutes per page. \clearpage
{
    \small
    \bibliographystyle{ieeenat_fullname}
    \bibliography{main, export}

\begin{thebibliography}{45}
\providecommand{\natexlab}[1]{#1}
\providecommand{\url}[1]{\texttt{#1}}
\expandafter\ifx\csname urlstyle\endcsname\relax
  \providecommand{\doi}[1]{doi: #1}\else
  \providecommand{\doi}{doi: \begingroup \urlstyle{rm}\Url}\fi

\bibitem[Anonymous et~al.(2022)Anonymous, community, and Branwen]{Anonymous2022}
Anonymous, Danbooru community, and Gwern Branwen.
\newblock Danbooru2021: A large-scale crowdsourced and tagged anime illustration dataset, 2022.

\bibitem[Carrillo et~al.(2023)Carrillo, Clément, Bugeau, and Simo-Serra]{Carrillo2023}
Hernan Carrillo, Michaël Clément, Aurélie Bugeau, and Edgar Simo-Serra.
\newblock Diffusart: Enhancing line art colorization with conditional diffusion models.
\newblock 2023.

\bibitem[Ci et~al.(2018)Ci, Ma, Wang, Li, and Luo]{Ci2018}
Yuanzheng Ci, Xinzhu Ma, Zhihui Wang, Haojie Li, and Zhongxuan Luo.
\newblock User-guided deep anime line art colorization with conditional adversarial networks.
\newblock 2018.

\bibitem[Esser et~al.(2021)Esser, Rombach, and Ommer]{Esser2021}
Patrick Esser, Robin Rombach, and Björn Ommer.
\newblock Taming transformers for high-resolution image synthesis.
\newblock 2021.

\bibitem[Furusawa et~al.(2017)Furusawa, Hiroshiba, Ogaki, and Odagiri]{Furusawa2017}
Chie Furusawa, Kazuyuki Hiroshiba, Keisuke Ogaki, and Yuri Odagiri.
\newblock Comicolorization: Semi-automatic manga colorization.
\newblock 2017.

\bibitem[Golyadkin and Makarov(2021)]{Golyadkin2021}
Maksim Golyadkin and Ilya Makarov.
\newblock Semi-automatic manga colorization using conditional adversarial networks.
\newblock 2021.

\bibitem[He et~al.(2016)He, Zhang, Ren, and Sun]{He2016}
Kaiming He, Xiangyu Zhang, Shaoqing Ren, and Jian Sun.
\newblock Deep residual learning for image recognition.
\newblock 2016.

\bibitem[Hensman and Aizawa(2018)]{Hensman2018}
Paulina Hensman and Kiyoharu Aizawa.
\newblock Cgan-based manga colorization using a single training image.
\newblock 2018.

\bibitem[Ho et~al.(2020)Ho, Jain, and Abbeel]{Ho2020}
Jonathan Ho, Ajay Jain, and Pieter Abbeel.
\newblock Denoising diffusion probabilistic models.
\newblock 2020.

\bibitem[Hu et~al.(2020)Hu, Shen, Albanie, Sun, and Wu]{Hu2020}
Jie Hu, Li Shen, Samuel Albanie, Gang Sun, and Enhua Wu.
\newblock Squeeze-and-excitation networks.
\newblock \emph{IEEE Transactions on Pattern Analysis and Machine Intelligence}, 42, 2020.

\bibitem[Im et~al.(2017)Im, Ahn, Memisevic, and Bengio]{Im2017}
Daniel~Jiwoong Im, Sungjin Ahn, Roland Memisevic, and Yoshua Bengio.
\newblock Denoising criterion for variational auto-encoding framework.
\newblock 2017.

\bibitem[Ishii et~al.(2009)Ishii, Kawamura, and Watanabe]{Ishii2009}
Daisuke Ishii, Kei Kawamura, and Hiroshi Watanabe.
\newblock A study on control parameters of frame decomposition method for comic images.
\newblock 2009.

\bibitem[Isola et~al.(2017)Isola, Zhu, Zhou, and Efros]{Isola2017}
Phillip Isola, Jun~Yan Zhu, Tinghui Zhou, and Alexei~A. Efros.
\newblock Image-to-image translation with conditional adversarial networks.
\newblock 2017.

\bibitem[Kim et~al.(2019)Kim, Jhoo, Park, and Yoo]{Kim2019}
Hyunsu Kim, Ho~Young Jhoo, Eunhyeok Park, and Sungjoo Yoo.
\newblock Tag2pix: Line art colorization using text tag with secat and changing loss.
\newblock 2019.

\bibitem[Kingma and Welling(2014)]{Kingma2014}
Diederik~P. Kingma and Max Welling.
\newblock Auto-encoding variational bayes.
\newblock 2014.

\bibitem[KumarSingh et~al.(2023)KumarSingh, Laddha, and James]{KumarSingh2023}
Nikhil KumarSingh, Nilay Laddha, and Joseph James.
\newblock An enhanced image colorization using modified generative adversarial networks with pix2pix method.
\newblock pages 1--8, 2023.

\bibitem[Larsen et~al.(2016)Larsen, Sønderby, Larochelle, and Winther]{Larsen2016}
Anders Boesen~Lindbo Larsen, Søren~Kaae Sønderby, Hugo Larochelle, and Ole Winther.
\newblock Autoencoding beyond pixels using a learned similarity metric.
\newblock 2016.

\bibitem[Li and Wand(2016)]{Li2016}
Chuan Li and Michael Wand.
\newblock Precomputed real-time texture synthesis with markovian generative adversarial networks.
\newblock 2016.

\bibitem[Lim and Ye(2017)]{Lim2017}
Jae~Hyun Lim and Jong~Chul Ye.
\newblock Geometric gan, 2017.

\bibitem[Liu et~al.(2023)Liu, Xing, Xie, Li, and Wong]{Liu2023}
Hanyuan Liu, Jinbo Xing, Minshan Xie, Chengze Li, and Tien-Tsin Wong.
\newblock Improved diffusion-based image colorization via piggybacked models, 2023.

\bibitem[Loshchilov and Hutter(2019)]{Loshchilov2019}
Ilya Loshchilov and Frank Hutter.
\newblock Decoupled weight decay regularization.
\newblock 2019.

\bibitem[mangadex(2023)]{mangadex2023}
mangadex.
\newblock mangadex, 2023.

\bibitem[Mirza and Osindero(2014)]{Mirza2014}
Mehdi Mirza and Simon Osindero.
\newblock Conditional generative adversarial nets, 2014.

\bibitem[Narang et~al.(2018)Narang, Diamos, Elsen, Micikevicius, Alben, Garcia, Ginsburg, Houston, Kuchaiev, Venkatesh, and Wu]{Narang2018}
Sharan Narang, Gregory Diamos, Erich Elsen, Paulius Micikevicius, Jonah Alben, David Garcia, Boris Ginsburg, Michael Houston, Oleksii Kuchaiev, Ganesh Venkatesh, and Hao Wu.
\newblock Mixed precision training.
\newblock 2018.

\bibitem[Nichol and Dhariwal(2021)]{Nichol2021}
Alex Nichol and Prafulla Dhariwal.
\newblock Improved denoising diffusion probabilistic models.
\newblock 2021.

\bibitem[Odena et~al.(2017)Odena, Olah, and Shlens]{Odena2017}
Augustus Odena, Christopher Olah, and Jonathon Shlens.
\newblock Conditional image synthesis with auxiliary classifier gans.
\newblock 2017.

\bibitem[pixiv(2023)]{pixiv2023}
pixiv.
\newblock pixiv, 2023.

\bibitem[Rodrigues et~al.(2022)Rodrigues, Clua, and Vitor]{Rodrigues2022}
Erick~Oliveira Rodrigues, Esteban Clua, and Giovani~Bernardes Vitor.
\newblock Line art colorization of fakemon using generative adversarial neural networks.
\newblock IEEE, 2022.

\bibitem[Rombach et~al.(2022)Rombach, Blattmann, Lorenz, Esser, and Ommer]{Rombach2022}
Robin Rombach, Andreas Blattmann, Dominik Lorenz, Patrick Esser, and Bjorn Ommer.
\newblock High-resolution image synthesis with latent diffusion models.
\newblock 2022.

\bibitem[Ronneberger et~al.(2015)Ronneberger, Fischer, and Brox]{Ronneberger2015}
Olaf Ronneberger, Philipp Fischer, and Thomas Brox.
\newblock U-net: Convolutional networks for biomedical image segmentation.
\newblock 2015.

\bibitem[Sato et~al.(2014)Sato, Matsui, Yamasaki, and Aizawa]{Sato2014}
Kazuhiro Sato, Yusuke Matsui, Toshihiko Yamasaki, and Kiyoharu Aizawa.
\newblock Reference-based manga colorization by graph correspondence using quadratic programming.
\newblock 2014.

\bibitem[Saxena(2022)]{Saxena2022}
Aarush Saxena.
\newblock An introduction to convolutional neural networks.
\newblock \emph{International Journal for Research in Applied Science and Engineering Technology}, 10, 2022.

\bibitem[Seo and Seo(2021)]{Seo2021}
Chang~Wook Seo and Yongduek Seo.
\newblock Seg2pix: Few shot training line art colorization with segmented image data.
\newblock \emph{Applied Sciences}, 11, 2021.

\bibitem[Shimizu et~al.(2021)Shimizu, Furuta, Ouyang, Taniguchi, Hinami, and Ishiwatari]{Shimizu2021}
Yugo Shimizu, Ryosuke Furuta, Delong Ouyang, Yukinobu Taniguchi, Ryota Hinami, and Shonosuke Ishiwatari.
\newblock Painting style-aware manga colorization based on generative adversarial networks.
\newblock 2021.

\bibitem[Silva et~al.(2019)Silva, Castro, Junior, and Marujo]{Silva2019}
Felipe~Coelho Silva, Paulo Andre Lima~De Castro, Helio~Ricardo Junior, and Ernesto~Cordeiro Marujo.
\newblock Mangan: Assisting colorization of manga characters concept art using conditional gan.
\newblock 2019.

\bibitem[Simonyan and Zisserman(2015)]{Simonyan2015}
Karen Simonyan and Andrew Zisserman.
\newblock Very deep convolutional networks for large-scale image recognition.
\newblock 2015.

\bibitem[Xie et~al.(2020)Xie, Li, Liu, and Wong]{Xie2020}
Minshan Xie, Chengze Li, Xueting Liu, and Tien~Tsin Wong.
\newblock Manga filling style conversion with screentone variational autoencoder.
\newblock \emph{ACM Transactions on Graphics}, 39, 2020.

\bibitem[Zeyen et~al.(2018)Zeyen, Post, Hagen, Ahrens, Rogers, and Bujack]{Zeyen2018}
Max Zeyen, Tobias Post, Hans Hagen, James Ahrens, David Rogers, and Roxana Bujack.
\newblock Color interpolation for non-euclidean color spaces.
\newblock pages 11--15, 2018.

\bibitem[Zhang et~al.(2018{\natexlab{a}})Zhang, Zuo, and Zhang]{Zhang2018_ffd}
Kai Zhang, Wangmeng Zuo, and Lei Zhang.
\newblock Ffdnet: Toward a fast and flexible solution for cnn-based image denoising.
\newblock \emph{IEEE Transactions on Image Processing}, 27, 2018{\natexlab{a}}.

\bibitem[Zhang et~al.(2018{\natexlab{b}})Zhang, Ji, Lin, and Liu]{Zhang2018style}
Lvmin Zhang, Yi Ji, Xin Lin, and Chunping Liu.
\newblock Style transfer for anime sketches with enhanced residual u-net and auxiliary classifier gan.
\newblock 2018{\natexlab{b}}.

\bibitem[Zhang et~al.(2021)Zhang, Li, Simo-Serra, Ji, Wong, and Liu]{Zhang2021}
Lvmin Zhang, Chengze Li, Edgar Simo-Serra, Yi Ji, Tien~Tsin Wong, and Chunping Liu.
\newblock User-guided line art flat filling with split filling mechanism.
\newblock 2021.

\bibitem[Zhang et~al.(2023)Zhang, Rao, and Agrawala]{Zhang2023}
Lvmin Zhang, Anyi Rao, and Maneesh Agrawala.
\newblock Adding conditional control to text-to-image diffusion models, 2023.

\bibitem[Zhang et~al.(2018{\natexlab{c}})Zhang, Li, Wong, Ji, and Liu]{Zhang2018}
Lv~Min Zhang, Chengze Li, Tien~Tsin Wong, Yi Ji, and Chun~Ping Liu.
\newblock Two-stage sketch colorization.
\newblock 2018{\natexlab{c}}.

\bibitem[Zhang et~al.(2018{\natexlab{d}})Zhang, Isola, Efros, Shechtman, and Wang]{Zhang2018_lpips}
Richard Zhang, Phillip Isola, Alexei~A. Efros, Eli Shechtman, and Oliver Wang.
\newblock The unreasonable effectiveness of deep features as a perceptual metric.
\newblock 2018{\natexlab{d}}.

\bibitem[Zhang et~al.(2009)Zhang, Chen, Zhang, Hu, and Martin]{Zhang2009}
Song~Hai Zhang, Tao Chen, Yi~Fei Zhang, Shi~Min Hu, and Ralph~R. Martin.
\newblock Vectorizing cartoon animations.
\newblock \emph{IEEE Transactions on Visualization and Computer Graphics}, 15, 2009.

\end{thebibliography}
}

\renewcommand{\thesection}{\Alph{section}}
\setcounter{section}{0}

\twocolumn[{
    \centering
    \Large
    \textbf{\thetitle}\\
    \vspace{0.5em}Supplementary Material \\
    \vspace{1.0em}

    \raggedright
    \normalsize

    \section{Qualitative ablation studies}
    \label{sec:a}

    \noindent\textbf{User form.} In our user study, we instructed participants to evaluate the visual quality of two different variants for each of the 26 samples in our questionnaire. The questionnaire was completed by 31 high school and first-year university students. The results of our study indicate a preference for the final post-processed results over the initial Style2PaintsV4.5 priors. It is worth noting that our primary focus both in the user study and the main paper was not on comparing our approach to the most competitive manga colorization method, manga-colorization-v2. The decision was made due to manga-colorization-v2 lacking support for color hinting, making it an unequal comparison. Furthermore, we aimed to avoid introducing bias in our study based on subjective opinions regarding color preferences.

    \begin{center}
        \centering
        \captionsetup{type=figure}
        \includegraphics[width=0.89\textwidth]{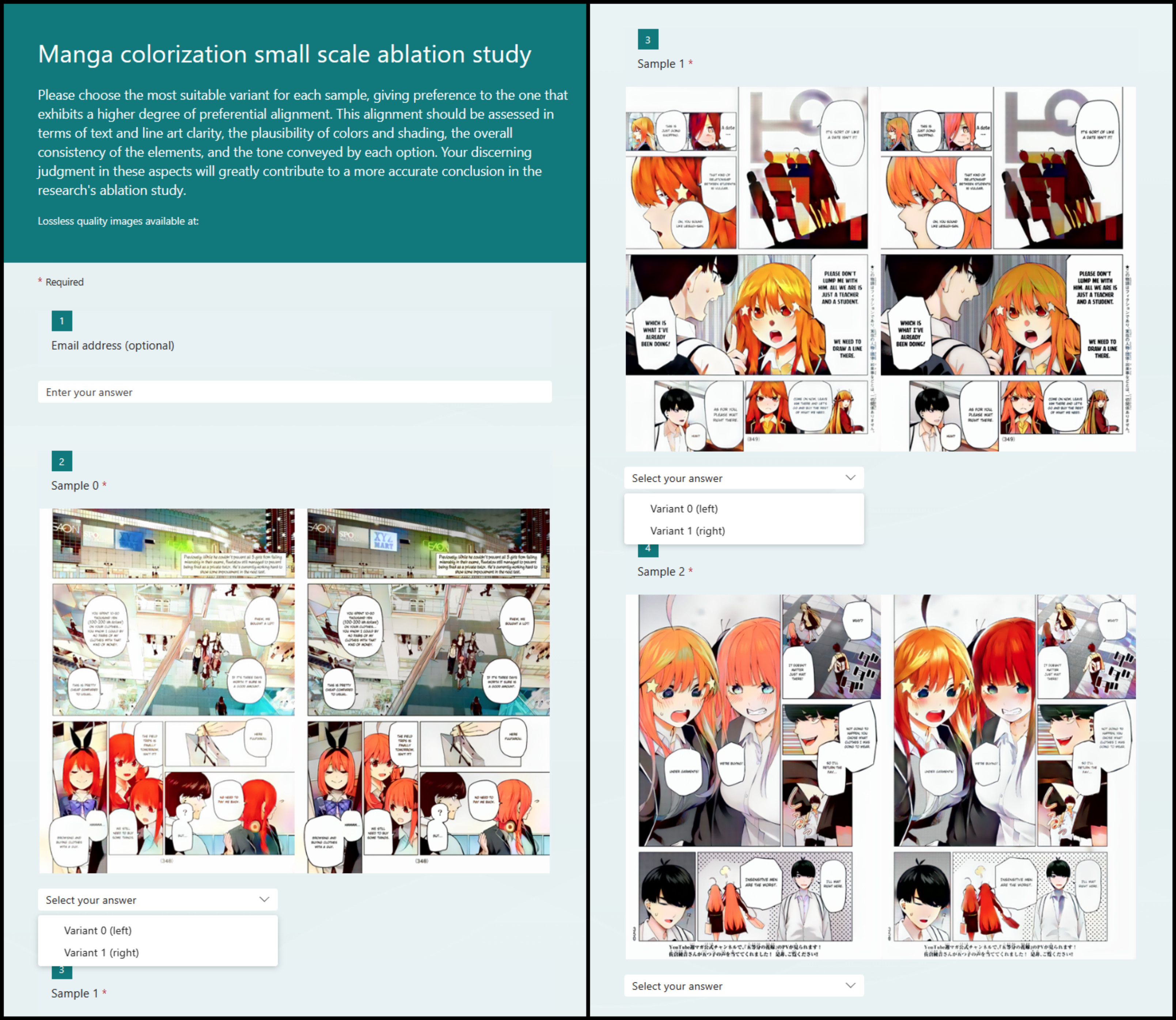}
        \captionof{figure}{\textbf{Example questionnaire} for the user qualitative ablation study}\label{fig:8}
    \end{center}
}]

\twocolumn[{
    \section{Additional colorization results}
    \label{sec:b}

    In this section, we present supplementary colorization results that have been generated using the process outlined in the main paper. These additional results were created using initial priors from Style2PaintsV4.5 using the default color palettes, without the use of reference images. Manual color hints were created based on the well-known character traits of each manga subject, ensuring that the colorization aligns with the specific characteristics and attributes of the characters within the manga.

    \begin{center}
        \centering
        \captionsetup{type=figure}
        \includegraphics[width=\textwidth]{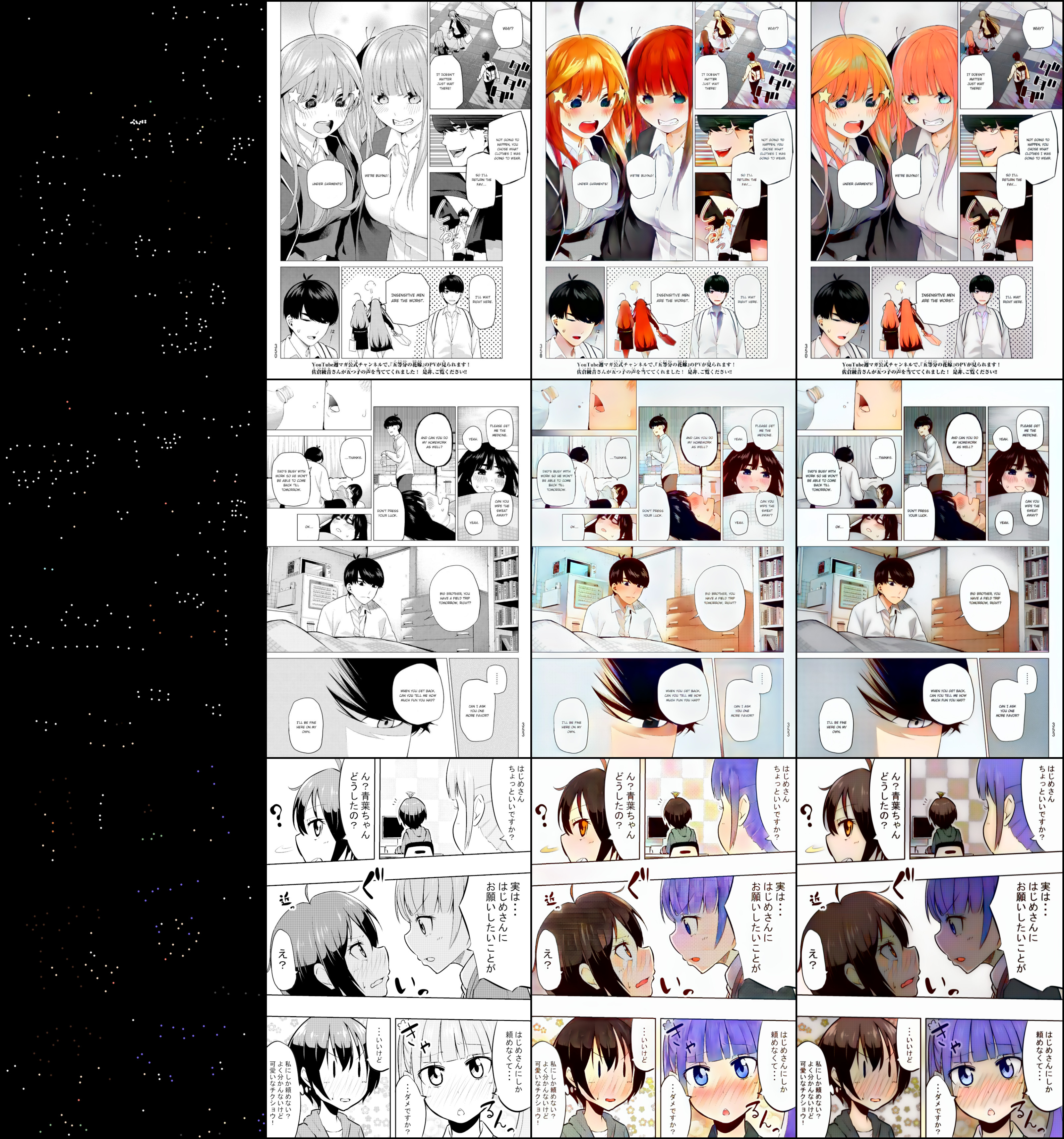}
        \scriptsize{\begin{tblr}{colsep = 0pt, colspec = {*{4}{X[c]}}}
            Color hints ($I_h$) & B\&W input ($I_{bw}$) & Rough-colored ($x_{col}$) & Ours ($\lambda_{a*b*} = 0.8$)
        \end{tblr}}
        \vspace*{-2.0em}
        \captionof{figure}{Additional colorization results are shown.}\label{fig:9}
    \end{center}
}]

\twocolumn[{
    \begin{center}
        \centering
        \captionsetup{type=figure}
        \includegraphics[width=\textwidth]{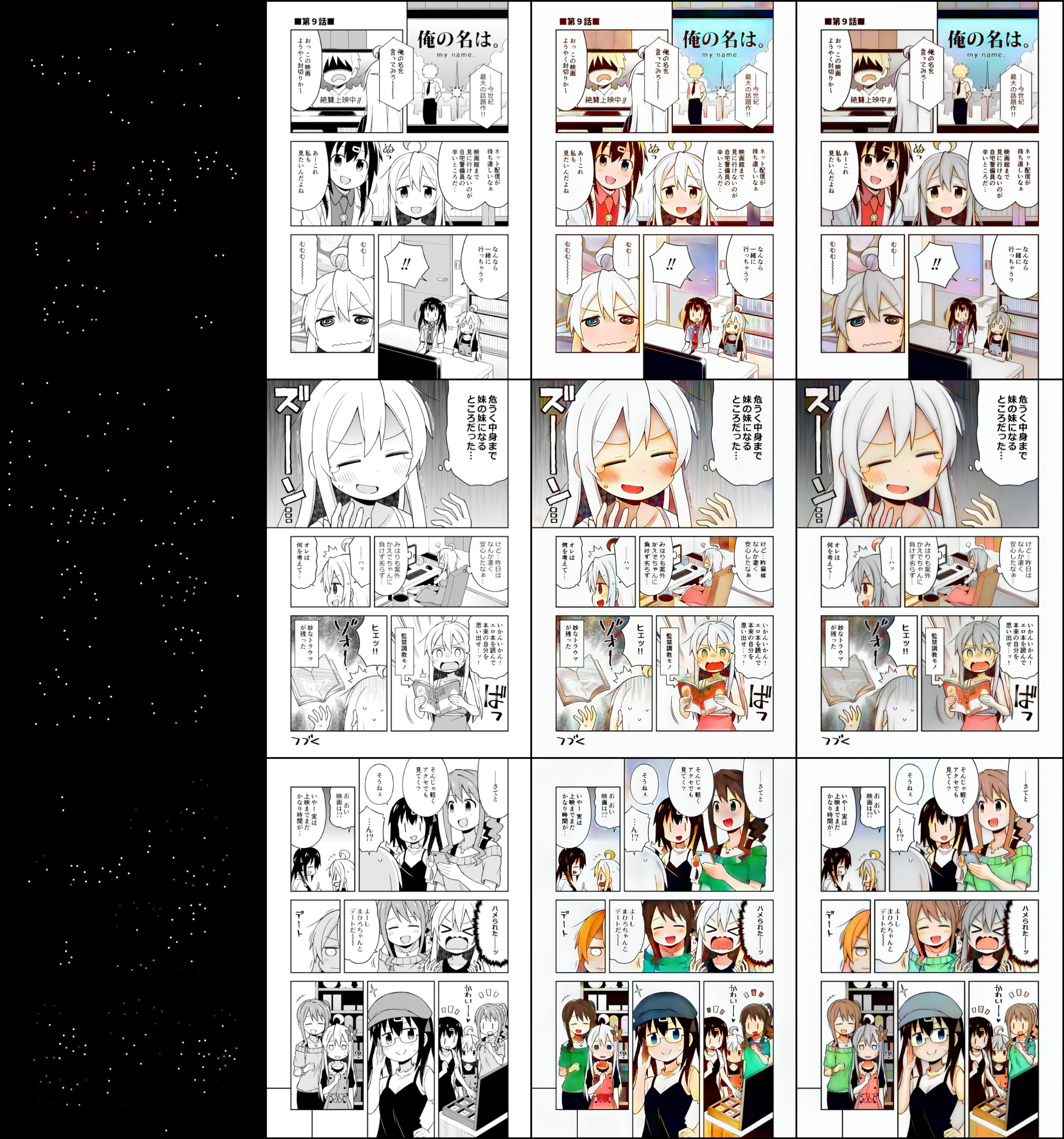}
        \scriptsize{\begin{tblr}{colsep = 0pt, colspec = {*{4}{X[c]}}}
            Color hints ($I_h$) & B\&W input ($I_{bw}$) & Rough-colored ($x_{col}$) & Ours ($\lambda_{a*b*} = 0.8$)
        \end{tblr}}
        \vspace*{-2.0em}
        \captionof{figure}{Additional colorization results are shown.}\label{fig:10}
    \end{center}
}]

\twocolumn[{
    \begin{center}
        \centering
        \captionsetup{type=figure}
        \includegraphics[width=\textwidth]{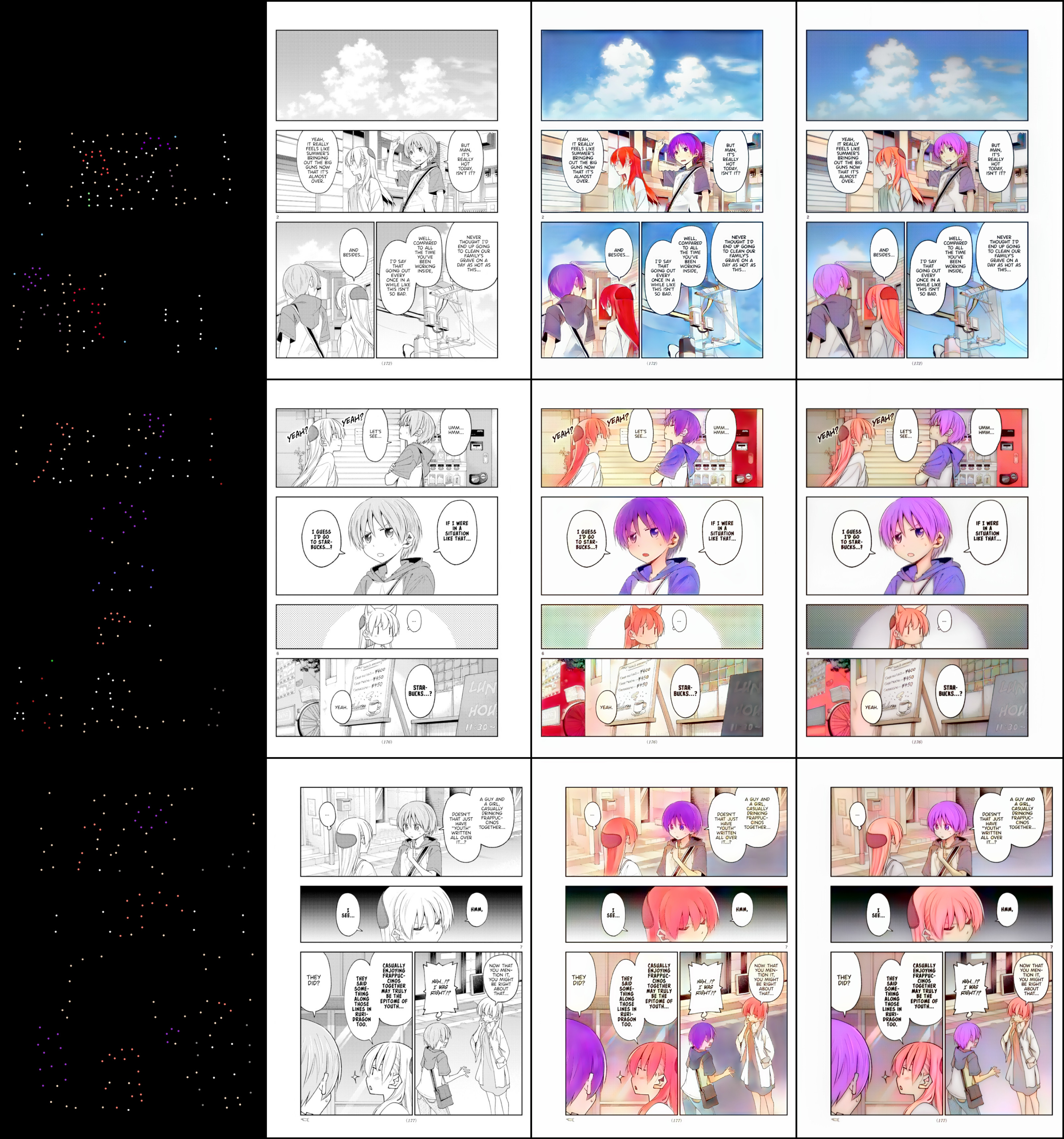}
        \scriptsize{\begin{tblr}{colsep = 0pt, colspec = {*{4}{X[c]}}}
            Color hints ($I_h$) & B\&W input ($I_{bw}$) & Rough-colored ($x_{col}$) & Ours ($\lambda_{a*b*} = 0.8$)
        \end{tblr}}
        \vspace*{-2.0em}
        \captionof{figure}{Additional colorization results are shown.}\label{fig:11}
    \end{center}
}]

\twocolumn[{
    \begin{center}
        \centering
        \captionsetup{type=figure}
        \includegraphics[width=\textwidth]{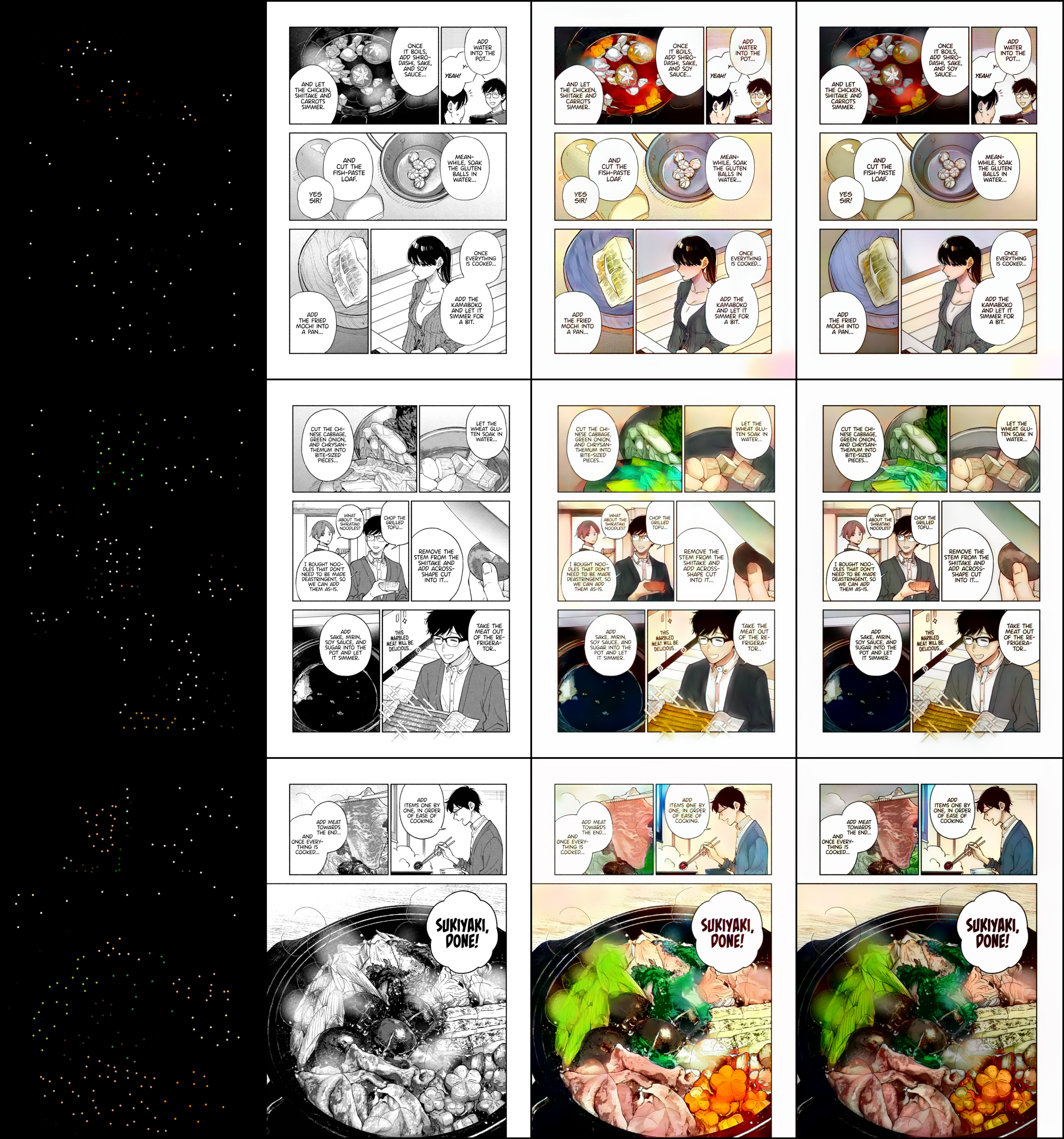}
        \scriptsize{\begin{tblr}{colsep = 0pt, colspec = {*{4}{X[c]}}}
            Color hints ($I_h$) & B\&W input ($I_{bw}$) & Rough-colored ($x_{col}$) & Ours ($\lambda_{a*b*} = 0.8$)
        \end{tblr}}
        \vspace*{-2.0em}
        \captionof{figure}{Additional colorization results are shown.}\label{fig:12}
    \end{center}
}]

\end{document}